# An Efficient Approach to Sparse Linear Discriminant Analysis

**Luis Francisco Sánchez Merchante**   luis-francisco.sanchez@utc.fr
**Yves Grandvalet**   yves.grandvalet@utc.fr
**Gérard Govaert**   gerard.govaert@utc.fr
Université de Technologie de Compiègne - CNRS UMR 7253 Heudiasyc, 60200 Compiègne, FRANCE

## Abstract

We present a novel approach to the formulation and the resolution of sparse Linear Discriminant Analysis (LDA). Our proposal, is based on penalized Optimal Scoring. It has an exact equivalence with penalized LDA, contrary to the multi-class approaches based on the regression of class indicator that have been proposed so far. Sparsity is obtained thanks to a group-Lasso penalty that selects the same features in all discriminant directions. Our experiments demonstrate that this approach generates extremely parsimonious models without compromising prediction performances. Besides prediction, the resulting sparse discriminant directions are also amenable to low-dimensional representations of data. Our algorithm is highly efficient for medium to large number of variables, and is thus particularly well suited to the analysis of gene expression data.

## 1. Introduction

Linear Discriminant Analysis (LDA) aims at finding the "best" separation of a set of observations into known classes. It is used for two main purposes: to classify future observations or to describe the essential differences between classes, either by providing a visual representation of data, or by revealing the combinations of features that discriminate between classes. An additional widely employed practice is to extract features by LDA, which is then used as a simple dimensionality reduction method taking into account the discriminant information.

LDA originates from the analysis of the within and between groups variances (Fisher, 1936), leading to the maximization of a separability criterion. It can be related to the estimation of the conditional probabilities of classes given the observations (without modelling their distribution), via a connection with the regression of class indicators. Finally, it can also be derived as a plug-in classifier under the assumption of normal distributions for the classes, with different means but a common variance matrix.

Sparse LDA refers here to formulations revealing discriminant directions that only involve a few variables. Besides cases where the sparsity of the true discriminants is assumed, such as most genetic analyses, sparse classification methods may be motivated by interpretability, robustness of the solution, or computational restraints for evaluation in prediction.

The most common approach to sparse LDA consists in performing variable selection in a separate step, before classification. Variable selection is then usually based on univariate statistics, which are fast and convenient to compute, but whose very partial view of the overall classification problem may lead to dramatic information loss. As a result, several approaches have been devised in the recent years to construct LDA with embedded feature selection capabilities.

Our Group-Lasso Optimal Scoring Solver (GLOSS) addresses a sparse LDA problem globally, through the regression approach of LDA. Our analysis formally relates GLOSS to Fisher's discriminant analysis, and also enables to derive variants, such that LDA assuming diagonal within-class covariance structure (Bickel & Levina, 2004). The group-Lasso selects the same features in all discriminant directions, which provide a more interpretable low-dimensional representation of data. Compared to the competing approaches, the models are extremely parsimonious without compromising prediction performances. Our algorithm efficiently processes medium to large number of variables, and is thus particularly well suited to the analysis of gene expression data.

This paper is organized as follows. Section 2 introduces the basic notations that are necessary for stating Fisher's discriminant problem. Section 3 reviews the main approaches that have been followed to perform sparse LDA via regression. We then derive a connection between sparse optimal scoring and sparse LDA in Section 4. The GLOSS algorithm is described in Section 5 and experimental results follow in Section 6, before our final concluding remarks of Section 7.

## 2. Primary Notations and Definitions

Vectors are denoted by lowercase letters in bold font and matrices by uppercase letters in bold font. Unless otherwise stated, vectors are column vectors and parentheses are used to build line vectors from comma-separated lists of scalars, or to build matrices from comma-separated lists of column vectors.

The data consist of a set of $n$ labeled examples, with observations $\mathbf{x}_i \in \mathbb{R}^p$ comprising $p$ features, and label $\mathbf{y}_i \in \{0,1\}^K$ indicating the exclusive assignment of observation $\mathbf{x}_i$ to one of the $K$ classes. It will be convenient to gather the observations in the $n \times p$ matrix $\mathbf{X} = (\mathbf{x}_1^\intercal, \ldots, \mathbf{x}_n^\intercal)^\intercal$ and the corresponding labels in the $n \times K$ matrix $\mathbf{Y} = (\mathbf{y}_1^\intercal, \ldots, \mathbf{y}_n^\intercal)^\intercal$.

The two main components of LDA are the within-class and between-class covariance matrices. We consider centered observations, so that the global sample mean is null, and $\mathbf{S} = n^{-1}\mathbf{X}^\intercal \mathbf{X}$ is the sample covariance matrix. The cardinality of class $k$ in the training sample is $n_k$, and the sample mean for class $k$ is $\hat{\boldsymbol{\mu}}_k$. The between-class sample covariance matrix is

$$\mathbf{S}_b = \frac{1}{n} \sum_{k=1}^{K} n_k \, \hat{\boldsymbol{\mu}}_k \hat{\boldsymbol{\mu}}_k^\intercal$$
$$= n^{-1} \mathbf{X}^\intercal \mathbf{P}_\mathbf{Y} \mathbf{X} \; ,$$

where $\mathbf{P}_\mathbf{Y} = \mathbf{Y}(\mathbf{Y}^\intercal \mathbf{Y})^{-1} \mathbf{Y}^\intercal$ projects onto the span of $\mathbf{Y}$. The within-class sample covariance matrix is

$$\mathbf{S}_w = \frac{1}{n} \sum_{k=1}^{K} \sum_{\{i : y_{ik}=1\}} (\mathbf{x}_i - \hat{\boldsymbol{\mu}}_k)(\mathbf{x}_i - \hat{\boldsymbol{\mu}}_k)^\intercal$$
$$= n^{-1} \mathbf{X}^\intercal (\mathbf{I}_n - \mathbf{P}_\mathbf{Y}) \mathbf{X} \; , \qquad (1)$$

where $\mathbf{I}_n$ is the identity matrix of size $n$.

The Fisher discriminant was originally defined in binary classification, as the linear projection that "best" separates the observations from each class, that is, maximizing the between-class variance relative to the within-class variance in the considered projection. This objective may also be stated as the maximization of between-class variance subject to unitary within-class variance, so as to define a unique maximizer.

When considering several projections, there are several equivalent formulations of LDA, which may differ once a penalization scheme is applied. Here, we use a definition based on subspace projection, where the discriminant directions maximally separate the class means subject to orthonormal constraints:

$$\mathbf{B}_{\text{LDA}} = \underset{\mathbf{B} \in \mathbb{R}^{p \times M}}{\operatorname{argmax}} \quad \operatorname{tr}(\mathbf{B}^\intercal \mathbf{S}_b \mathbf{B}) \\ \text{s. t.} \quad \mathbf{B}^\intercal \mathbf{S}_w \mathbf{B} = \mathbf{I}_M \; , \qquad (2)$$

where $\mathbf{B}_{\text{LDA}} = (\boldsymbol{\beta}_1, \ldots, \boldsymbol{\beta}_M)$ gathers the $M \leq K-1$ discriminant directions, and $\operatorname{tr}(\cdot)$ is the trace operator. This definition was chosen here for its succinctness, but it is incomplete with regard to the usual textbook LDA procedure where the discriminant directions are computed iteratively so as to maximally separate the class means subject to orthonormality constraints: here, the discriminant directions are defined up to an orthonormal transformation. Note that this is a notational problem and that we can recover such an ordering in the GLOSS procedure proposed here.

## 3. Sparse LDA

LDA is often used as a data reduction technique, where the $K-1$ discriminant directions summarize the $p$ original variables. However, all variables intervene in the definition of these discriminant directions, and this trait may be troublesome in some applications.

Several modifications of LDA have been proposed to generate sparse discriminant directions. They can be categorized according to the LDA formulation that provides the basis to the sparsity inducing extension, that is, either Fisher's Discriminant Analysis (variance-based), Gaussian mixture model (based on the assumption of the normality of classes) or regression-based. We briefly review here the approaches belonging to the last category to introduce the method proposed in this paper.

Fisher (1936) introduced what is now known as *Fisher's linear discriminant analysis* in his analysis of the famous iris dataset, and discussed its analogy with the linear regression of the scaled class indicators. This route was further developed, for more than two classes, by Breiman & Ihaka (1984) as an inspiration for a non-linear extension of discriminant analysis using additive models. They named their approach *optimal scaling*, for it optimizes the scaling of the indicators of classes together with the discriminant functions. Their approach later disseminated under the name *optimal scoring* (OS) by Hastie et al. (1994), who proposed several extensions of LDA, either aiming at constructing more flexible discriminants (Hastie & Tibshirani, 1996) or more conservative ones (Hastie et al., 1995).

Several sparse LDA have been derived using sparsity-inducing penalties on the OS regression problem (Leng, 2008; Grosenick et al., 2008; Clemmensen et al., 2011). These proposals are motivated by the equivalence between penalized OS and penalized LDA, but since sparsity-inducing penalties are non-quadratic, they fall beyond the realm of the equivalence stated by Hastie et al. (1995). Until now, no rigorous link was derived between sparse OS and sparse LDA.

In this paper, we demonstrate that the equivalence between penalized OS and penalized LDA is preserved for the non-quadratic Lasso penalty for binary classification. However, the connection collapses in the general multi-class setting. We thus propose GLOSS, based on the group-Lasso, that preserves the equivalence between sparse OS and sparse LDA in the general multi-class situation.

## 4. From Sparse OS to Sparse LDA

We relate here a sparse OS problem, penalized by the group-Lasso, with a sparse LDA problem. Our derivation uses a variational formulation of the group-Lasso to generalize the equivalence drawn by Hastie et al. (1995) for quadratic penalties.

### 4.1. A Variational Form of the Group-Lasso

Quadratic variational forms of the Lasso and group-Lasso have been proposed shortly after the original Lasso paper of Tibshirani (1996), as a means to address optimization issues, but also as an inspiration for generalizing the Lasso penalty (Grandvalet, 1998; Grandvalet & Canu, 1999). The algorithms based on these quadratic variational forms iteratively reweight a quadratic penalty. They are now often outperformed by more efficient strategies (Bach et al., 2012).

#### 4.1.1. A Quadratic Variational Form

We now present a handy convex quadratic variational form of the group-Lasso. Let $\mathbf{B} \in \mathbb{R}^{p \times M}$ be a matrix composed of row vectors $\boldsymbol{\beta}^j \in \mathbb{R}^M$, $\mathbf{B} = \left(\boldsymbol{\beta}^{1\mathsf{T}}, \ldots, \boldsymbol{\beta}^{p\mathsf{T}}\right)^\mathsf{T}$. We consider the following problem:

$$\min_{\boldsymbol{\tau} \in \mathbb{R}^p} \min_{\mathbf{B} \in \mathbb{R}^{p \times M}} \quad J(\mathbf{B}) + \lambda \sum_{j=1}^{p} w_j^2 \frac{\left\|\boldsymbol{\beta}^j\right\|_2^2}{\tau_j} \quad \text{(3a)}$$

$$\text{s. t.} \quad \sum_j \tau_j - \sum_j w_j \left\|\boldsymbol{\beta}^j\right\|_2 \leq 0 \quad \text{(3b)}$$

$$\tau_j \geq 0 \, , \, j = 1, \ldots, p \, . \quad \text{(3c)}$$

where $w_j$ are predefined weights. Here and in what follows, $b/\tau$ is defined by continuation at zero as $b/0 = \infty$ if $b \neq 0$ and $0/0 = 0$. Note that variants of (3) have been proposed elsewhere (see e.g. Grandvalet & Canu, 1999; Bach et al., 2012, and references therein).

**Lemma 1.** *The quadratic penalty in $\boldsymbol{\beta}^j$ in (3) acts as the group-Lasso penalty $\lambda \sum_{j=1}^{p} w_j \left\|\boldsymbol{\beta}^j\right\|_2$.*

This equivalence is crucial to the derivation of the link between sparse OS and sparse LDA; it furthermore suggests a convenient implementation. We sketch below some properties that are instrumental in the implementation of the active-set described in Section 5.

#### 4.1.2. Useful Properties

The first property states that the quadratic formulation is convex when $J$ is convex, thus providing an easy control of optimality and convergence.

**Lemma 2.** *If $J$ is convex, Problem (3) is convex.*

In what follows, $J$ will be a convex quadratic (hence smooth) function, in which case a necessary and sufficient optimality condition is that zero belongs to the subdifferential of the objective function. This condition results in an equality for the "active" non-zero vectors $\boldsymbol{\beta}^j$, and an inequality for the other ones, which both provide essential building blocks of our algorithm.

**Lemma 3.** *Problem (3) admits at least one solution, which is unique if $J$ is strictly convex. All critical points $\mathbf{B}$ of the objective function verifying the following conditions are global minima.*

$$\forall j \in \mathcal{S}, \quad \frac{\partial J(\mathbf{B})}{\partial \boldsymbol{\beta}^j} + \lambda w_j \left\|\boldsymbol{\beta}^j\right\|_2^{-1} \boldsymbol{\beta}^j = 0 \, , \quad \text{(4a)}$$

$$\forall j \in \mathcal{S}^c, \quad \left\|\frac{\partial J(\mathbf{B})}{\partial \boldsymbol{\beta}^j}\right\|_2 \leq \lambda \, . \quad \text{(4b)}$$

*where $\mathcal{S} \subseteq \{1, \ldots, p\}$ denotes the set of non-zero row vectors $\boldsymbol{\beta}^j$ and $\mathcal{S}^c(\mathbf{B})$ is its complement.*

Lemma 3 provides a simple appraisal of the support of the solution, which would not be as easily handled with the direct analysis of the variational problem (3).

### 4.2. Penalized Optimal Scoring

In binary classification, the regression of (scaled) class indicators enables to recover the LDA discriminant direction. For more than two classes, this approach is impaired by the masking effect (Hastie et al., 1994), where the scores assigned to a class situated between two other ones may never dominate. Optimal scoring (OS) circumvents the problem by assigning "optimal scores" to the classes.

Hastie et al. (1995) proposed to incorporate a smoothness prior on the discriminant directions in the OS problem through a positive-definite penalty matrix $\mathbf{\Omega}$, leading to a problem expressed in compact form as

$$\min_{\mathbf{\Theta}, \mathbf{B}} \|\mathbf{Y\Theta} - \mathbf{XB}\|_F^2 + \lambda \operatorname{tr}(\mathbf{B}^\intercal \mathbf{\Omega B}) \quad (5)$$
$$\text{s. t. } \mathbf{\Theta}^\intercal \mathbf{Y}^\intercal \mathbf{Y\Theta} = \mathbf{I}_{K-1} ,$$

where $\mathbf{\Theta}$ are the class scores, $\mathbf{B}$ the regression coefficients, and $\|\cdot\|_F$ is the Frobenius norm.

Hastie et al. (1995) proved that, under the assumption that $\mathbf{Y}^\intercal \mathbf{Y}$ and $\mathbf{X}^\intercal \mathbf{X} + \lambda \mathbf{\Omega}$ are full rank (which is fulfilled when there are no empty class and $\mathbf{\Omega}$ is positive definite), Problem (5) is equivalent to a penalized LDA problem, where the sample within-class covariance $\mathbf{S}_w$ defined in (1) is replaced in (2) by the "shrunken" estimate

$$\hat{\mathbf{\Sigma}}_w = n^{-1} \left( \mathbf{X}^\intercal \left( \mathbf{I}_n - \mathbf{P_Y} \right) \mathbf{X} + \lambda \mathbf{\Omega} \right)$$
$$= \mathbf{S}_w + n^{-1} \lambda \mathbf{\Omega} . \quad (6)$$

Note that $\hat{\mathbf{\Sigma}}_w$ has larger eigenvalues than $\mathbf{S}_w$; shrinkage refers here to the bias of $\hat{\mathbf{\Sigma}}_w$ towards $\mathbf{\Omega}$. Linear discriminant classifiers are invariant to the "size" of $\hat{\mathbf{\Sigma}}_w$, that is, not modified by an overall scaling of $\hat{\mathbf{\Sigma}}_w$.

The equivalence states that the solutions in $\mathbf{B}$ to the OS problem can be mapped to the solutions of the corresponding penalized LDA problem. The parameters of this mapping are furthermore computed when solving the OS optimization problem.

Though non-convex, the OS problem is readily solved by a decomposition in $\mathbf{\Theta}$ and $\mathbf{B}$: the optimal $\mathbf{B}^\star$ does not intervene in the optimality conditions with respect to $\mathbf{\Theta}$ and the optimization with respect to $\mathbf{B}$ is obtained in a closed form as a linear combination of the optimal scores $\mathbf{\Theta}^\star$ (Hastie et al., 1995). The algorithm may seem a bit tortuous considering the properties mentioned above, as it proceeds in four steps:

1. initialize $\mathbf{\Theta}$ to $\mathbf{\Theta}^0$ such that $\mathbf{\Theta}^{0\intercal} \mathbf{Y}^\intercal \mathbf{Y\Theta}^0 = \mathbf{I}_{K-1}$;

2. compute $\mathbf{B} = (\mathbf{X}^\intercal \mathbf{X} + \lambda \mathbf{\Omega})^{-1} \mathbf{X}^\intercal \mathbf{Y\Theta}^0$;

3. set $\mathbf{\Theta}^\star$ to be the $K-1$ leading eigenvectors of $\mathbf{Y}^\intercal \mathbf{X} (\mathbf{X}^\intercal \mathbf{X} + \lambda \mathbf{\Omega})^{-1} \mathbf{X}^\intercal \mathbf{Y}$;

4. compute the optimal regression coefficients

$$\mathbf{B}^\star = (\mathbf{X}^\intercal \mathbf{X} + \lambda \mathbf{\Omega})^{-1} \mathbf{X}^\intercal \mathbf{Y\Theta}^\star . \quad (7)$$

Defining $\mathbf{\Theta}^0$ in Step 1, instead of using directly $\mathbf{\Theta}^\star$ as expressed in Step 3, drastically reduces the computational burden of the eigenanalysis: the latter is performed on $\mathbf{\Theta}^{0\intercal} \mathbf{Y}^\intercal \mathbf{X} (\mathbf{X}^\intercal \mathbf{X} + \lambda \mathbf{\Omega})^{-1} \mathbf{X}^\intercal \mathbf{Y\Theta}^0$, which is computed as $\mathbf{\Theta}^{0\intercal} \mathbf{Y}^\intercal \mathbf{XB}$, thus avoiding a costly matrix inversion. Finally, note that $(\mathbf{\Theta}^\star, \mathbf{B}^\star)$ are uniquely defined up to sign swaps and column permutations, and that all critical points are global optima.

### 4.3. Group-Lasso OS as Penalized LDA

We now have all the necessary ingredients to introduce our Group-Lasso Optimal Scoring Solver for performing sparse LDA.

**Proposition 1.** *The Group-Lasso OS problem*

$$\mathbf{B}^\star = \operatorname*{argmin}_{\mathbf{B} \in \mathbb{R}^{p \times M}} \min_{\mathbf{\Theta} \in \mathbb{R}^{K \times M}} \frac{1}{2} \|\mathbf{Y\Theta} - \mathbf{XB}\|_F^2 + \lambda \sum_{j=1}^p \|\boldsymbol{\beta}^j\|_2$$
$$\text{s. t. } \mathbf{\Theta}^\intercal \mathbf{Y}^\intercal \mathbf{Y\Theta} = \mathbf{I}_M ,$$

*with $M = K - 1$, is equivalent to the penalized LDA problem*

$$\mathbf{B}_{\text{LDA}} = \operatorname*{argmax}_{\mathbf{B} \in \mathbb{R}^{p \times M}} \operatorname{tr}(\mathbf{B}^\intercal \mathbf{S}_b \mathbf{B})$$
$$\text{s. t. } \mathbf{B}^\intercal (\mathbf{S}_w + n^{-1} \lambda \mathbf{\Omega}) \mathbf{B} = \mathbf{I}_M ,$$

*that is, $\mathbf{B}_{\text{LDA}} = \mathbf{B}^\star \operatorname{diag}\left( (\alpha_k^{-1}(1 - \alpha_k^2)^{-1/2}) \right)$, where $\alpha_k \in (0, 1)$ is the kth leading eigenvalue of*

$$\mathbf{M} = n^{-1} \mathbf{Y}^\intercal \mathbf{X} (\mathbf{X}^\intercal \mathbf{X} + \lambda \mathbf{\Omega})^{-1} \mathbf{X}^\intercal \mathbf{Y} ,$$

*with $\mathbf{\Omega} = \operatorname{diag}\left( \|\boldsymbol{\beta}^{1\star}\|_2^{-1}, \ldots, \|\boldsymbol{\beta}^{p\star}\|_2^{-1} \right)$, using again the convention that $\|\boldsymbol{\beta}^{j\star}\|_2$ null implies that the jth row of $\mathbf{B}_{\text{LDA}}$ is null.*

*Proof.* The proof simply consists in applying the result of Hastie et al. (1995), which holds for quadratic penalties, to the quadratic variational form of the group-lasso. □

The proposition applies in particular to the Lasso-based OS approaches to sparse LDA (Grosenick et al., 2008; Clemmensen et al., 2011) for $K = 2$, that is, for binary classification or more generally for a single discriminant direction. Note however that it leads to a slightly different decision rule if the decision threshold is chosen *a priori* according to the Gaussian assumption for the features. For more than one discriminant direction, the equivalence does not hold any more, since the Lasso penalty does not result in an equivalent quadratic penalty in the simple form $\operatorname{tr}(\mathbf{B}^\intercal \mathbf{\Omega B})$.

**Algorithm 1** Adaptively Penalized Optimal Scoring
**Input:** $\mathbf{X}$, $\mathbf{Y}$, $\mathbf{B}$, $\lambda$
  Initialize: $\mathcal{S} \leftarrow \{j \in \{1,\ldots,p\} : \|\boldsymbol{\beta}^j\|_2 > 0\}$,
  $\boldsymbol{\Theta}^0 : \boldsymbol{\Theta}^{0\mathsf{T}}\mathbf{Y}^{\mathsf{T}}\mathbf{Y}\boldsymbol{\Theta}^0 = \mathbf{I}_{K-1}$, convergence $\leftarrow$ **false**
  **repeat**
    // Step 1: solve (3) in $\mathbf{B}$ assuming $\mathcal{S}$ optimal
    **repeat**
      $\boldsymbol{\Omega} \leftarrow \mathrm{diag}(\boldsymbol{\omega}_\mathcal{S})$, with $\omega_j \leftarrow \|\boldsymbol{\beta}^j\|_2^{-1}$
      $\mathbf{B}^\mathcal{S} \leftarrow (\mathbf{X}_{\boldsymbol{\cdot}\mathcal{S}}^\mathsf{T}\mathbf{X}_{\boldsymbol{\cdot}\mathcal{S}} + \lambda\boldsymbol{\Omega})^{-1}\mathbf{X}_{\boldsymbol{\cdot}\mathcal{S}}^\mathsf{T}\mathbf{Y}\boldsymbol{\Theta}^0$
    **until** conditions (4) hold for all $j \in \mathcal{S}$
    // Step 2: identify inactivated variables
    **for** $\{j \in \mathcal{S} : \|\boldsymbol{\beta}^j\|_2 = 0\}$ **do**
      **if** optimality condition (4b) holds **then**
        $\mathcal{S} \leftarrow \mathcal{S} \setminus \{j\}$
      **end if**
    **end for**
    // Step 3: check for optimality of set $\mathcal{S}$
    $j^\star \leftarrow \underset{j\in\mathcal{S}^c}{\mathrm{argmax}} \|\partial J/\partial\boldsymbol{\beta}^j\|_2$
    **if** $\|\partial J/\partial\boldsymbol{\beta}^{j^\star}\|_2 < \lambda$ **then**
      convergence $\leftarrow$ **true** // $\mathbf{B}$ is optimal
    **else**
      $\mathcal{S} \leftarrow \mathcal{S} \cup \{j^\star\}$
    **end if**
  **until** convergence
  $(\mathbf{s}, \mathbf{V}) \leftarrow$ eigenanalyze$(\boldsymbol{\Theta}^{0\mathsf{T}}\mathbf{Y}^\mathsf{T}\mathbf{X}_{\boldsymbol{\cdot}\mathcal{S}}\mathbf{B})$, that is,
      $\boldsymbol{\Theta}^{0\mathsf{T}}\mathbf{Y}^\mathsf{T}\mathbf{X}_{\boldsymbol{\cdot}\mathcal{S}}\mathbf{B}\mathbf{v}_k = s_k\mathbf{v}_k$ $k = 1,\ldots K-1$
  $\boldsymbol{\Theta}^\star \leftarrow \boldsymbol{\Theta}^0\mathbf{V}$, $\mathbf{B}^\star \leftarrow \mathbf{B}\mathbf{V}$, $\alpha_k^\star \leftarrow n^{-1/2}s_k^{1/2}$
**Output:** $\boldsymbol{\Theta}^\star$, $\mathbf{B}^\star$, $\boldsymbol{\alpha}^\star$

## 5. GLOSS Algorithm

The efficient approaches developed for the Lasso take advantage of the sparsity of the solution by solving a series of small linear systems, whose sizes are incrementally increased/decreased (Osborne et al., 2000). This approach was pursued for the group-Lasso (Roth & Fischer, 2008) in its standard formulation. We adapt this algorithmic framework to the variational form (3), with $J(\mathbf{B}) = 1/2\|\mathbf{Y}\boldsymbol{\Theta} - \mathbf{X}\mathbf{B}\|_F^2$.

The algorithm starts from a sparse initial guess, say $\mathbf{B} = \mathbf{0}$, thus defining the set $\mathcal{S}$ of "active" variables, currently identified as non-zero. Then, it iterates the three steps summarized in Algorithm 1.

### 5.1. OS Regression Coefficients Updates

Step 1 of Algorithm 1 updates the coefficient matrix $\mathbf{B}$ within the current active set $\mathcal{S}$. The quadratic variational form of the problem suggests a blockwise optimization strategy consisting in solving $(K-1)$ independent card$(\mathcal{S})$-dimensional problems instead of a single $(K-1) \times$ card$(\mathcal{S})$-dimensional problem. The interaction between the $(K-1)$ problems is relegated to the common adaptive quadratic penalty $\boldsymbol{\Omega}$. This decomposition is especially attractive as we then solve $(K-1)$ similar systems:

$$(\mathbf{X}_{\boldsymbol{\cdot}\mathcal{S}}^\mathsf{T}\mathbf{X}_{\boldsymbol{\cdot}\mathcal{S}} + \lambda\boldsymbol{\Omega})\boldsymbol{\beta}_k = \mathbf{X}_{\boldsymbol{\cdot}\mathcal{S}}^\mathsf{T}\mathbf{Y}\boldsymbol{\theta}_k^0 ,$$

where $\mathbf{X}_{\boldsymbol{\cdot}\mathcal{S}}$ denotes the columns of $\mathbf{X}$ indexed by $\mathcal{S}$ and $\boldsymbol{\beta}_k$ and $\boldsymbol{\theta}_k^0$ denote the $k$th column of $\mathbf{B}$ and $\boldsymbol{\Theta}^0$ respectively. These linear systems only differ in the right-hand-side term, so that a single Cholesky decomposition is necessary to solve all systems, whereas a blockwise Newton-Raphson method based on the standard group-Lasso formulation would result in different "penalties" $\boldsymbol{\Omega}$ for each system.

### 5.2. Solution Path

Finally, note that our default strategy is to compute a series of solutions along the regularization path, defined by a series of penalties $\lambda_1 = \lambda_{\max} > \cdots > \lambda_t > \cdots > \lambda_T = \lambda_{\min} \geq 0$ such that $\mathbf{B}^\star(\lambda_{\max}) = \mathbf{0}$, that is $\lambda_{\max} = \max_{j\in\{1,\ldots,p\}}\|\mathbf{X}_j^\mathsf{T}\mathbf{Y}\boldsymbol{\Theta}^0\|_2$, where $\mathbf{X}_j$ is the $j$th column of $\mathbf{X}$. Then, we regularly decrease the penalty $\lambda_{t+1} = \lambda_t/2$ and use a warm-start strategy, where the feasible initial guess for $\mathbf{B}(\lambda_{t+1})$ is initialized with $\mathbf{B}(\lambda_t)$. The final penalty parameter $\lambda_{\min}$ is specified in the optimization process when the maximum number of desired active variables is attained (by default the minimum of $n$ and $p$).

### 5.3. Diagonal LDA Variant

We motivated the group-Lasso penalty by sparsity requisites, but robustness considerations could also drive its usage, since LDA is known to be unstable when the number of examples is small compared to the number of variables. In this context, LDA has been experimentally observed to benefit from unrealistic assumptions on the form of the estimated within-class covariance matrix. Indeed, the diagonal approximation that ignores correlations between genes may lead to better classification in microarray analysis. Bickel & Levina (2004) shown that this crude approximation provides a classifier with best worst-case performances than the LDA decision rule in small sample size regimes, even if variables are correlated.

The equivalence proof between penalized OS and penalized LDA of Hastie et al. (1995) reveals that quadratic penalties in the OS problem are equivalent to penalties on the within-class covariance matrix in the LDA formulation. This proof suggests a slight variant of penalized OS (5) corresponding to penalized LDA with diagonal within-class covariance ma-trix, where the least square problems

$$\|\mathbf{Y}\boldsymbol{\Theta} - \mathbf{XB}\|_F^2 = \operatorname{tr}(\boldsymbol{\Theta}^\mathsf{T}\mathbf{Y}^\mathsf{T}\mathbf{Y}\boldsymbol{\Theta} - 2\boldsymbol{\Theta}^\mathsf{T}\mathbf{XB} + n\mathbf{B}^\mathsf{T}\mathbf{SB})$$

are replaced by

$$\operatorname{tr}(\boldsymbol{\Theta}^\mathsf{T}\mathbf{Y}^\mathsf{T}\mathbf{Y}\boldsymbol{\Theta} - 2\boldsymbol{\Theta}^\mathsf{T}\mathbf{XB} + n\mathbf{B}^\mathsf{T}(\mathbf{S}_b + \operatorname{diag}(\mathbf{S}_w))\mathbf{B})$$

Note that this variant requires $\operatorname{diag}(\mathbf{S}_w) + \boldsymbol{\Omega} - \mathbf{S}$ to be positive definite.

## 6. Experimental Results

This section presents some experimental results comparing our proposed algorithm, GLOSS, to two other sparse linear classifiers recently proposed to perform sparse LDA, namely the Penalized LDA (PLDA) of Witten & Tibshirani (2011), which applies a Lasso penalty in Fisher's LDA framework, and the Sparse Linear Discriminant Analysis (SLDA) of Clemmensen et al. (2011), which applies an elastic net penalty to the OS problem. The latter was used with a single tuning coefficient, without any quadratic term in the penalty, so that the elastic net reduces to the Lasso. The two competitors where implemented with the code publicly available from the authors in R and MATLAB respectively (Witten, 2011; Clemmensen, 2008). All results have been computed using the same training, validation and test sets. Note that they differ significantly from the ones of Witten & Tibshirani (2011) in Simulation 4 for which there was a typo in their paper.

### 6.1. Simulated Data

We first compare the three techniques in the simulation study of Witten & Tibshirani (2011), which considers four setups with 1200 examples equally distributed between classes. They are split in a training set of size $n = 100$, a validation set of size 100, and a test set of size 1000. We are in the small sample regime, with $p = 500$ variables, out of which 100 differ between classes. Independent variables are generated for all simulations except for simulation 2 where they are slightly correlated. In simulation 2 and 3, classes are optimally separated by a single projection of the original variables, while the two other scenarios require three discriminant directions. The Bayes' error was estimated to be respectively 1.7%, 6.7%, 7.3% and 30.0%. We follow all the other details of the simulation protocol of Witten & Tibshirani (2011).

Note that this protocol is detrimental to GLOSS as each relevant variable only affects a single class mean out of $K$. The setup is favorable to PLDA in the sense that most within-class covariance matrix are diagonal. We thus tested the diagonal GLOSS variant discussed in Section 5.3.

Table 1. Simulation results: averages, with standard errors computed over 25 repetitions, of the test error rate, the number of selected variables and the number of discriminant directions selected on the validation set.

|  | Err. (%) | # Var. | # Dir. |
|---|---|---|---|
| *Sim. 1: $K=4$, mean shift, ind. features* | | | |
| PLDA | 12.6 (0.1) | 411.7 (3.7) | 3.0 (0.0) |
| SLDA | 31.9 (0.1) | 228.0 (0.2) | 3.0 (0.0) |
| GLOSS | 19.9 (0.1) | 106.4 (1.3) | 3.0 (0.0) |
| GLOSS-D | 11.2 (0.1) | 251.1 (4.1) | 3.0 (0.0) |
| *Sim. 2: $K=2$, mean shift, dependent features* | | | |
| PLDA | 9.0 (0.4) | 337.6 (5.7) | 1.0 (0.0) |
| SLDA | 19.3 (0.1) | 99.0 (0.0) | 1.0 (0.0) |
| GLOSS | 15.4 (0.1) | 39.8 (0.8) | 1.0 (0.0) |
| GLOSS-D | 9.0 (0.0) | 203.5 (4.0) | 1.0 (0.0) |
| *Sim. 3: $K=4$, 1D mean shift, ind. features* | | | |
| PLDA | 13.8 (0.6) | 161.5 (3.7) | 1.0 (0.0) |
| SLDA | 57.8 (0.2) | 152.6 (2.0) | 1.9 (0.0) |
| GLOSS | 31.2 (0.1) | 123.8 (1.8) | 1.0 (0.0) |
| GLOSS-D | 18.5 (0.1) | 357.5 (2.8) | 1.0 (0.0) |
| *Sim. 4: $K=4$, mean shift, ind. features* | | | |
| PLDA | 60.3 (0.1) | 336.0 (5.8) | 3.0 (0.0) |
| SLDA | 65.9 (0.1) | 208.8 (1.6) | 2.7 (0.0) |
| GLOSS | 60.7 (0.2) | 74.3 (2.2) | 2.7 (0.0) |
| GLOSS-D | 58.8 (0.1) | 162.7 (4.9) | 2.9 (0.0) |

The results are summarized in Table 1. Overall, the best predictions are performed by PLDA and GLOS-D that both benefit of the knowledge of the true within-class covariance structure. Then, among SLDA and GLOSS that both ignore this structure, our proposal has a clear edge. The error rates are far away from the Bayes' error rates, but the sample size is small with regard to the number of relevant variables. Regarding sparsity, the clear overall winner is GLOSS, followed far away by SLDA, which is the only method that do not succeed in uncovering a low-dimensional representation in Simulation 3. The adequacy of the selected features was assessed by the *True Positive Rate* (TPR) and the *False Positive Rate* (FPR). PLDA has the best TPR but a terrible FPR, except in simulation 3 where it dominates all the other methods. GLOSS has by far the best FPR with overall TPR slightly below SLDA.

### 6.2. Gene Expression Data

We now compare GLOSS to PLDA and SLDA on three genomic datasets. The *Nakayama* dataset contains 105 examples of 22,283 gene expressions for categorizing 10 soft tissue tumors. It was reduced to the 86 examples belonging to the 5 dominant categories (Witten & Tibshirani, 2011). The *Ramaswamy* dataset contains 198 examples of 16,063 gene expressions for cat-egorizing 14 classes of cancer. Finally, the *Sun* dataset contains 180 examples of 54,613 gene expressions for categorizing 4 classes of tumors.

Table 2. Gene expression results: averages over 10 training/test sets splits, with standard deviations, of the test error rates and the number of selected variables.

|  | Err. (%) | # Var. |
|---|---|---|
| *Nakayama*: $n = 86$, $p = 22{,}283$, $K = 5$ | | |
| PLDA | 20.95 (1.3) | 10,478.7 (2,116.3) |
| SLDA | 25.71 (1.7) | 252.5 (3.1) |
| GLOSS | 20.48 (1.4) | 129.0 (18.6) |
| *Ramaswamy*: $n = 198$, $p = 16{,}063$, $K = 14$ | | |
| PLDA | 38.36 (6.0) | 14,873.5 (720.3) |
| SLDA | — | — |
| GLOSS | 20.61 (6.9) | 372.4 (122.1) |
| *Sun*: $n = 180$, $p = 54{,}613$, $K = 4$ | | |
| PLDA | 33.78 (5.9) | 21,634.8 (7,443.2) |
| SLDA | 36.22 (6.5) | 384.4 (16.5) |
| GLOSS | 31.77 (4.5) | 93.0 (93.6) |

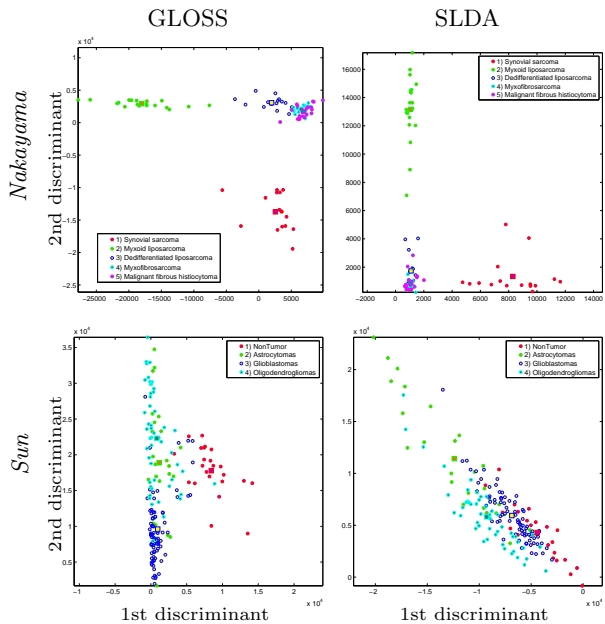

Figure 1. 2D-representations of Nakayama and Sun datasets based on the two first discriminant vectors provided by GLOSS and SLDA. The big squares represent

Each dataset was split into a training set and a test set with respectively 75% and 25% of the examples. The tuning parameter is performed by 10-fold cross-validation and the test performances are then evaluated. The process is repeated 10 times, with random choices of training and test set split.

We present the test error rates and the number of selected variables in Table 2. The three methods have comparable prediction performances on the *Nakayama* and *Sun* data, but GLOSS performs better on the *Ramaswamy* data, where the SparseLDA package failed to return a solution, due to numerical problems in the LARS-EN implementation. Regarding the number of selected variables, GLOSS is again much sparser than its competitors.

Finally, Figure 1 displays the projection of the observations for the *Nakayama* and *Sun* datasets in the first canonical planes estimated by GLOSS and SLDA. For the *Nakayama* dataset, groups 1 and 2 are well-separated from the other ones in both representations, but GLOSS is more discriminant in the meta-cluster gathering groups 3 to 5. For the *Sun* dataset, SLDA suffers from a high colinearity of its first canonical variables that renders the second one almost non-informative. As a result, group 1 is better separated in the first canonical plane with GLOSS.

## 7. Conclusions and Further Works

We described GLOSS, an efficient algorithm that performs sparse LDA based on the regression of class indicators. Our proposal is equivalent to a penalized LDA problem. This is up to our knowledge the first approach that enjoys this property in the multi-class setting. This relationship is also amenable to accommodate interesting constraints on the equivalent penalized LDA problem, such as imposing a diagonal structure of the within-class covariance matrix.

Computationally, GLOSS is based on an efficient active set strategy that is amenable to the processing of problems with a large number of variables. The inner optimization problem decouples the $p \times (K-1)$-dimensional problem into $(K-1)$ independent $p$-dimensional problems. The interaction between the $(K-1)$ problems is relegated to the computation of the common adaptive quadratic penalty. The algorithm presented here is highly efficient in medium to high dimensional setups, which makes it a good candidate for the analysis of gene expression data.

Our experimental results confirm the relevance of the approach, which behaves well compared to its competitors, either regarding its prediction abilities or its interpretability (sparsity). Employing the same features in all discriminant directions enables to generate models that are globally extremely parsimonious, with good prediction abilities. The resulting sparse discriminant directions also allow for visual inspection of data from the low-dimensional representations that can be produced.

The approach has many potential extensions that have not yet been implemented. A first line of development is to consider a broader class of penalties. For example, plain quadratic penalties can also be added to the group-penalty to encode priors about the within-class covariance structure, in the spirit of the Penalized Discriminant Analysis of Hastie et al. (1995). Also, besides the group-Lasso, our framework can be customized to any penalty that is uniformly spread within groups, and many composite or hierarchical penalties that have been proposed for structured data meet this condition.

## Acknowledgments

This research was supported by the PASCAL2 Network of Excellence, the European ICT FP7 (grant 247022 - MASH), and the French National Research Agency ANR (grant ClasSel 08-EMER-002).